\documentclass[11pt]{article}
\usepackage{amsthm}

\usepackage[final]{acl}
\usepackage{adjustbox}
\usepackage{array}
\usepackage{threeparttable}
\usepackage{tabularx}
\usepackage{graphicx}
\usepackage{subcaption}
\usepackage{times}
\usepackage{latexsym}
\usepackage[T1]{fontenc}
\usepackage[utf8]{inputenc}
\usepackage{microtype}
\usepackage{inconsolata}
\usepackage{tikz}

\usetikzlibrary{positioning, calc, decorations.pathreplacing}

\usepackage{amsmath}
\usepackage{amsmath}
\usepackage{amssymb}
\usepackage{booktabs}
\usepackage{pifont}

\usepackage{graphicx}
\usepackage{tikz}
\usepackage{pgfplots}
\pgfplotsset{compat=1.18}
\usepgfplotslibrary{groupplots}

\usepackage{booktabs}
\usepackage{multirow}
\usepackage{makecell}

\usepackage{algpseudocode}

\usepackage{algorithm}
\newtheorem{proposition}{Proposition}

\title{SOS-LoRA: Static Orthogonal-Subspace Low-Rank Adaptation with Fixed Multi-Scale Scaling}

\author{
    Yupeng Chang$^{1}$ \quad Yuan Wu$^{1}$\footnotemark[1] \quad Yi Chang$^{1,2,3}$ \\
    $^{1}$School of Artificial Intelligence, Jilin University \\
    $^{2}$Engineering Research Center of Knowledge-Driven Human-Machine Intelligence, MOE, China \\
    $^{3}$International Center of Future Science, Jilin University \\
    changyp23@mails.jlu.edu.cn, \{yuanwu, yichang\}@jlu.edu.cn
}

\begin{document}
\maketitle

\renewcommand{\thefootnote}{\fnsymbol{footnote}}
\footnotetext[1]{Corresponding author}

\begin{abstract}
Low-Rank Adaptation (LoRA) is a widely used parameter-efficient fine-tuning (PEFT) method for large language models. Under a fixed rank budget, LoRA parameterizes each adapted weight through a single low-dimensional input-side pathway, which may couple heterogeneous behaviors through shared input directions and induce interference during optimization. We propose \textbf{Static Orthogonal Subspace LoRA} (SOS-LoRA), a drop-in extension that reparameterizes a rank-$r_{\text{tot}}$ update as a sum of $K$ \emph{static} (always-on, non-routed) low-rank experts. SOS-LoRA (i) decomposes the total rank across experts, (ii) applies a \emph{fixed} multi-scale scaling scheme to encourage scale-separated optimization dynamics, and (iii) promotes diverse input-side directions via cross-expert orthogonal initialization and a lightweight regularizer. SOS-LoRA remains fully mergeable, adding no inference-time parameters or latency after merging. Experiments on reasoning and knowledge-intensive benchmarks (Llama 2/3), encoder-based NLU (GLUE), and math reasoning (GSM8K/MATH) show consistent gains over matched-budget LoRA baselines and recent variants. Code is available at \url{https://github.com/llm172/sos-lora}.
\end{abstract}

\section{Introduction}
\label{sec:introduction}

Parameter-Efficient Fine-Tuning (PEFT) is widely used to adapt large language models (LLMs)~\citep{touvron2023llama,chang2024survey, han2024parameter} when full fine-tuning is computationally and operationally expensive. Among PEFT strategies, Low-Rank Adaptation (LoRA)~\citep{hu2022lora} is a standard choice due to its simplicity, strong empirical performance, and deployment-friendly mergeability. LoRA is motivated by the observation that task-specific weight updates often exhibit low effective dimensionality, consistent with evidence on low intrinsic dimension in fine-tuning~\citep{aghajanyan2020intrinsic}. Concretely, LoRA freezes pretrained weights $W_0$ and learns a rank-$r_{\text{tot}}$ update
$\Delta W = \frac{\alpha}{r_{\text{tot}}}AB$,
where $A \in \mathbb{R}^{m \times r_{\text{tot}}}$ and $B \in \mathbb{R}^{r_{\text{tot}} \times n}$, yielding the adapted weight $W_0 + \Delta W$. Since the update is additive and linear, it can be merged into $W_0$ after training, preserving the backbone architecture and inference efficiency~\citep{hu2022lora}.

Despite its success, standard LoRA assigns each adapted matrix a \emph{single} low-rank update pathway. This raises a natural question: under a fixed rank budget, can one shared low-dimensional pathway adequately support the diverse behaviors required by modern LLMs? In the standard factorization, the adaptation signal depends on the input only through a single projection $XA$, i.e., a single set of input-side directions shared by all adaptation effects. Consequently, heterogeneous behaviors may be forced to share and compete for the same input-side directions under a fixed rank budget, which can induce interference and optimization coupling. Increasing the rank can partially alleviate this issue, but it also increases trainable parameters and still provides no explicit mechanism for structured adaptation under a fixed budget.

In this work, we mitigate this coupling by \emph{structuring} the low-rank update into multiple expert components without increasing the total rank budget. We propose \textbf{Static Orthogonal Subspace LoRA} (\textbf{SOS-LoRA}), a drop-in replacement for standard LoRA that reparameterizes a rank-$r_{\text{tot}}$ update as a sum of $K$ \emph{static} (always-on, non-routed) low-rank experts. Each expert uses its own input-side directions, is assigned a fixed scale, and is explicitly encouraged to remain directionally distinct from other experts. Importantly, SOS-LoRA does not enlarge the hypothesis class beyond rank-$r_{\text{tot}}$ updates; rather, it introduces an optimization-oriented inductive bias that promotes directionally diverse adaptation under the same low-rank budget.

SOS-LoRA instantiates this inductive bias through three components. First, a \emph{parallel expert decomposition} splits the total rank budget into $K$ experts of rank $r' = r_{\text{tot}}/K$, matching the parameter budget of standard LoRA at the same $r_{\text{tot}}$. Second, a \emph{fixed multi-scale scaling} scheme assigns each expert a static scale spanning small-to-large magnitudes; this scales the raw backpropagated gradients entering each expert and encourages scale-separated learning dynamics. Third, \emph{orthogonal input-side diversification} combines cross-expert orthogonal initialization (with zero initial update) and an orthogonality regularizer that penalizes cross-expert correlations in input-side directions.

Crucially, SOS-LoRA preserves LoRA's key deployment property: after training, all expert updates can be merged into the pretrained weights as a single effective update matrix. Deployment therefore uses the same backbone architecture and per-token FLOPs as the base model, incurring no additional inference latency after merging~\citep{hu2022lora}.

Our contributions are threefold:
\begin{itemize}
    \item We identify a key limitation of standard LoRA under a fixed rank budget: a single shared input-side projection can couple heterogeneous behaviors through a common low-dimensional update pathway, which can induce interference and optimization coupling.
    \item We propose \textbf{SOS-LoRA}, which reparameterizes a rank-$r_{\text{tot}}$ LoRA update as a sum of $K$ \emph{static} low-rank experts with fixed multi-scale scaling and explicit cross-expert diversification of input-side directions, while preserving LoRA's mergeability and inference efficiency.
    \item We empirically evaluate SOS-LoRA on reasoning and knowledge-intensive benchmarks, encoder-based NLU tasks (GLUE), and math reasoning (GSM8K/MATH), showing consistent improvements over matched-budget LoRA baselines and recent LoRA variants under a unified training and evaluation pipeline.
\end{itemize}

\begin{figure*}[t]
    \centering
    \includegraphics[width=.86\textwidth]{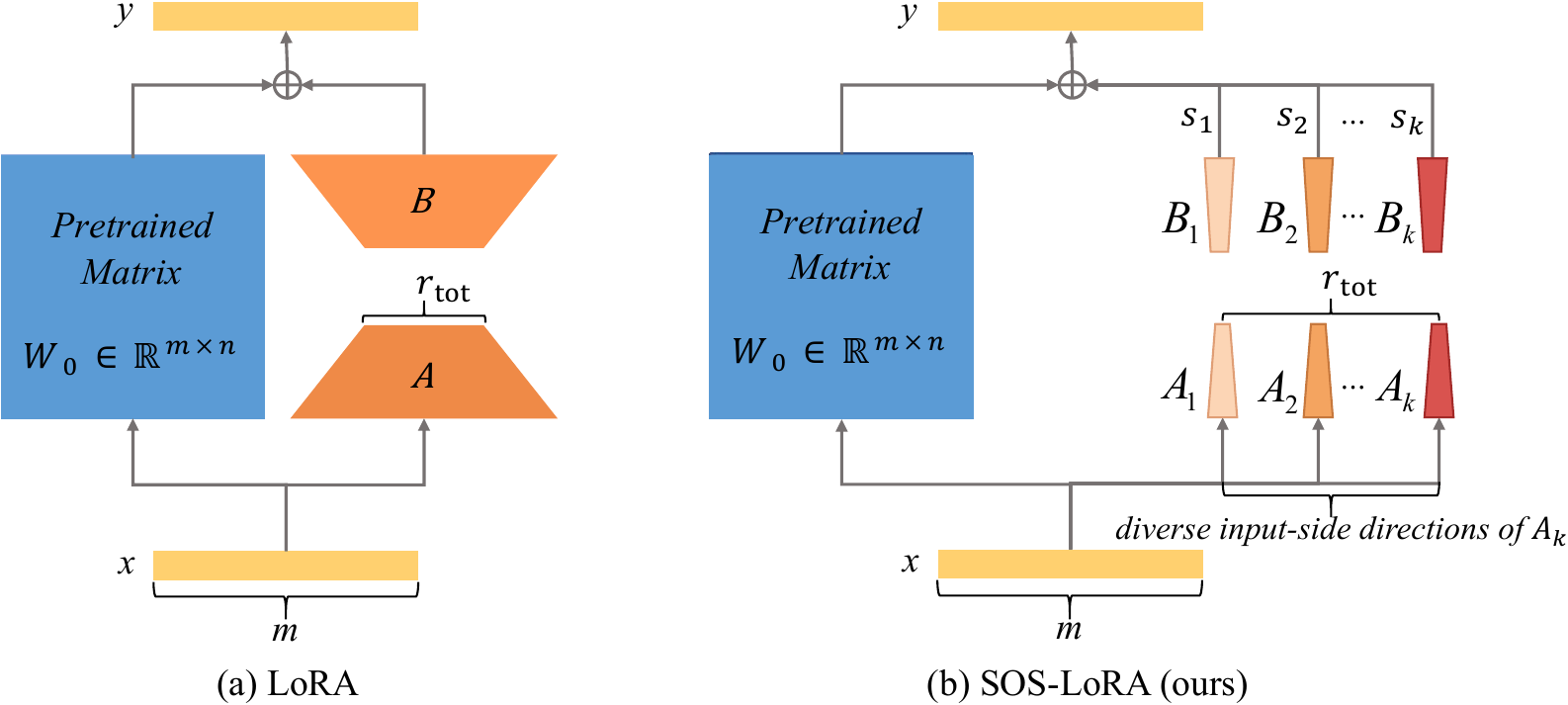}
\caption{Overview of Static Orthogonal Subspace LoRA (SOS-LoRA) with row-vector convention $y=xW_0$, $W_0\in\mathbb{R}^{m\times n}$, $x\in\mathbb{R}^{1\times m}$, and $y\in\mathbb{R}^{1\times n}$. 
(a) LoRA freezes $W_0$ and learns a single rank-$r_{\text{tot}}$ update $\Delta W=\frac{\alpha}{r_{\text{tot}}}AB$ with $A\in\mathbb{R}^{m\times r_{\text{tot}}}$ and $B\in\mathbb{R}^{r_{\text{tot}}\times n}$, yielding $y=x(W_0+\Delta W)$; under this parameterization, the adapter depends on the input only through the shared projection $xA$. 
(b) SOS-LoRA reparameterizes the same total rank budget as $K$ \emph{static} (always-on, non-routed) rank-$r'$ experts $\{(A_k,B_k)\}_{k=1}^K$ with $r'=r_{\text{tot}}/K$, combined by fixed multi-scale scales $\{s_k\}$ and encouraged to use diverse input-side directions via an orthogonality regularizer on $\{A_k\}$. The merged update is $\Delta W=\sum_{k=1}^K s_kA_kB_k$, which can be absorbed into $W_0$ with no additional inference-time parameters or latency after merging.}
    \label{fig:sos_lora_overview}
\end{figure*}

\section{Related Work}
\label{sec:related_work}

Parameter-Efficient Fine-Tuning (PEFT) adapts large pretrained models by updating only a small subset of parameters~\citep{han2024parameter}. Representative approaches include adding lightweight modules such as Adapters~\citep{houlsby2019parameter}, prompt-based tuning that optimizes continuous prompts or prefixes~\citep{lester2021power,li2021prefix}, and low-dimensional reparameterizations of weight updates. Among these, Low-Rank Adaptation (LoRA)~\citep{hu2022lora} and its subsequent extensions~\citep{chang2026balora,chang2025lora} freeze pretrained weights and learn efficient low-rank updates, making LoRA-based methods a widely used PEFT family.

A growing body of work extends LoRA along multiple axes. Optimization- and training-focused variants aim to improve convergence or update quality (e.g., LoRA+~\citep{hayou2024lora+}, LoRA-GA~\citep{wang2024loraga}, LoRA-Pro~\citep{wang2024loraprolowrankadaptersproperly}), while orthogonality-based methods encourage less interfering low-rank updates (e.g., O-LoRA~\citep{wang2023orthogonal}, primarily studied in continual learning). Other directions explore multiple LoRA experts through token-dependent routing (e.g., MixLoRA~\citep{li2024mixlora}), combine multiple \emph{trained} LoRAs via learnable gating (e.g., MoLE~\citep{wu2024mixture}), or introduce multi-scale designs across layers (e.g., MSPLoRA~\citep{zhao2025msplora}). Related ideas on enhancing specialization or expressiveness have also been explored in nonlinear LoRA designs and neuron-level mixture frameworks~\citep{dong2025aurora,dong2026neureasoner}. HD-PiSSA~\citep{wang2025hd} also employs orthogonal adapter subspaces, but initializes them from disjoint singular components of pretrained weights across devices and applies their induced updates to shared weights while keeping adapter factors fixed, aiming to increase effective update rank. In contrast, \textbf{SOS-LoRA} jointly trains $K$ \emph{static} (always-on, non-routed) low-rank experts \emph{within each adapted weight} under a fixed total rank budget, combining random QR initialization, fixed multi-scale scaling, and cross-expert input-side orthogonality regularization to bias optimization while preserving LoRA's mergeability and expressivity bound.

\section{Methodology}
\label{sec:methodology}

We propose \textbf{Static Orthogonal Subspace LoRA (SOS-LoRA)}, a parameter-efficient fine-tuning (PEFT) method that aims to improve \emph{effective} adaptation under a fixed low-rank budget. SOS-LoRA does so by: (i) decomposing a single rank-$r_{\text{tot}}$ update into multiple \emph{static} (always-on, non-routed) low-rank experts; (ii) using a \emph{pre-specified} multi-scale scaling schedule (optionally with mild \emph{input-independent} calibration) to promote stable, scale-separated optimization dynamics; and (iii) explicitly encouraging \emph{cross-expert diversity of input-side directions} through initialization and a lightweight regularizer. Like standard LoRA~\citep{hu2022lora}, SOS-LoRA is \emph{mergeable} into linear layers: after fine-tuning, the learned low-rank updates can be merged into the base weights, yielding zero additional inference-time parameters and latency after merging. Figure~\ref{fig:sos_lora_overview} provides an overview of the standard LoRA parameterization and the SOS-LoRA design.

\subsection{Problem Formulation: Input-Side Coupling Under a Fixed Rank Budget}
\label{ssec:problem_formulation}

Let $W_0 \in \mathbb{R}^{m \times n}$ be a pre-trained weight matrix in a Transformer linear layer. Let $X \in \mathbb{R}^{N \times m}$ denote the input activations for $N$ token instances (flattened across batch and sequence), and let $Y \in \mathbb{R}^{N \times n}$ denote the corresponding outputs. Standard Low-Rank Adaptation (LoRA)~\citep{hu2022lora} freezes $W_0$ and parameterizes an additive update $\Delta W$ with rank at most $r_{\text{tot}}$:
\begin{equation}
    Y = X(W_0 + \Delta W)
      = X\!\left(W_0 + \frac{\alpha}{r_{\text{tot}}} A B\right),
    \label{eq:lora}
\end{equation}
where $A \in \mathbb{R}^{m \times r_{\text{tot}}}$, $B \in \mathbb{R}^{r_{\text{tot}} \times n}$, and $\alpha$ is a scaling scalar.

\paragraph{Shared input-side pathway.}
Eq.~\eqref{eq:lora} implies that the LoRA path depends on the input only through a single low-dimensional projection. Writing $Z = X A \in \mathbb{R}^{N \times r_{\text{tot}}}$, we have $XAB = ZB$, hence the LoRA contribution depends on $X$ only via $Z$. In particular, for any $X_1, X_2$ such that $X_1A = X_2A$, the LoRA path yields identical outputs $X_1AB = X_2AB$. Equivalently, for a single token row $x \in \mathbb{R}^{1 \times m}$, the LoRA path depends on $x$ only through $xA$ (or $A^\top x^\top$). Under a fixed rank budget, this means that all input-sensitive adaptation effects are mediated through the same shared rank-$r_{\text{tot}}$ input projection.

\paragraph{Optimization coupling under a fixed rank budget.}
Increasing $r_{\text{tot}}$ increases the maximum rank of $\Delta W$ but also increases the number of trainable parameters. Under a \emph{fixed} $r_{\text{tot}}$, a single factorization $AB$ can couple heterogeneous adaptation signals by forcing them to share and compete for the same input-side directions in $A$. SOS-LoRA introduces a structured parameterization and explicit cross-expert direction regularization to bias optimization toward more diverse input directions, while preserving the same overall rank budget.

\subsection{SOS-LoRA: Architecture Design}
\label{ssec:sos_arch}

SOS-LoRA replaces a single rank-$r_{\text{tot}}$ adapter with $K$ parallel rank-$r'$ experts, where $r' = r_{\text{tot}}/K$ (assuming $K$ divides $r_{\text{tot}}$). The design has three components: (i) parallel expert decomposition, (ii) multi-scale scaling, and (iii) input-side diversification via orthogonal initialization and regularization.

\paragraph{Expressivity equivalence (hypothesis class).}
SOS-LoRA does \emph{not} enlarge the set of representable rank-$r_{\text{tot}}$ updates compared to standard LoRA; instead, it provides an optimization-oriented inductive bias through scale-separated optimization and cross-expert direction diversity. Formally:

\begin{proposition}[Expressivity equivalence]
\label{prop:equiv}
Let $\{(A_k,B_k)\}_{k=1}^K$ with $A_k \in \mathbb{R}^{m \times r'}$ and $B_k \in \mathbb{R}^{r' \times n}$, and let $s_k$ be scalars. Define the column concatenation
$A_{\mathrm{cat}} = [A_1,\ldots,A_K] \in \mathbb{R}^{m \times r_{\text{tot}}}$
and the row concatenation
$B_{\mathrm{cat}} = [s_1B_1;\ldots;s_KB_K] \in \mathbb{R}^{r_{\text{tot}} \times n}$.
Then
$\sum_{k=1}^K s_k A_k B_k = A_{\mathrm{cat}} B_{\mathrm{cat}}$,
so SOS-LoRA represents the same set of updates with rank at most $r_{\text{tot}}$ as a single LoRA adapter (up to rank deficiency).
\end{proposition}

\subsubsection{Parallel Experts and Multi-Scale Scaling}
\label{ssec:parallel_experts}

We denote expert parameters by $\{(A_k, B_k)\}_{k=1}^K$, where
$A_k \in \mathbb{R}^{m \times r'}$ and $B_k \in \mathbb{R}^{r' \times n}$.
The SOS-LoRA forward pass is
\begin{equation}
    Y_{\text{SOS}} = XW_0 + \sum_{k=1}^{K} s_k \, X A_k B_k,
    \label{eq:sos_forward}
\end{equation}
where $s_k > 0$ controls the contribution of expert $k$.
Here, ``static'' means that (i) all experts are active for all tokens, (ii) there is no token-dependent routing, and (iii) the decomposition introduces no conditional computation.

\paragraph{Fixed multi-scale base scaling.}
We initialize expert scales as
\begin{equation}
    \label{eq:multi_scale}
    \begin{split}
        s_k &= \frac{\alpha}{r_{\text{tot}}} \cdot \frac{\gamma_k}{\bar{\gamma}}, \\
        \gamma_k &= 1 + \frac{k-1}{\max(1, K-1)}(\gamma_{\max}-1), \\
        \bar{\gamma} &= \frac{1}{K}\sum_{k=1}^K \gamma_k,
    \end{split}
\end{equation}
so that the \emph{average} scaling matches standard LoRA,
$\frac{1}{K}\sum_{k=1}^K s_k = \alpha/r_{\text{tot}}$,
while experts receive linearly spaced relative scales $\gamma_k \in [1,\gamma_{\max}]$.
Unless stated otherwise, we use $\gamma_{\max}=2.5$.

\paragraph{Scale-separated gradient signals.}
Let $\Delta = \partial \mathcal{L} / \partial Y_{\text{SOS}} \in \mathbb{R}^{N \times n}$ denote the upstream gradient for loss $\mathcal{L}$. From Eq.~\eqref{eq:sos_forward}, the per-expert gradients satisfy
\begin{equation}
    \frac{\partial \mathcal{L}}{\partial B_k}
    = s_k \, (X A_k)^\top \Delta,
    \qquad
    \frac{\partial \mathcal{L}}{\partial A_k}
    = s_k \, X^\top (\Delta B_k^\top),
    \label{eq:grad_scale}
\end{equation}
so $s_k$ linearly scales the raw backpropagated gradients entering expert $k$.
With adaptive optimizers, the resulting parameter \emph{updates} are not strictly linear in $s_k$ due to normalization effects; nevertheless, fixed multi-scale scaling provides a simple and stable mechanism to bias experts toward different update magnitudes.

\paragraph{Optional input-independent calibration (training-time only).}
In practice, we optionally allow a mild \emph{input-independent} reweighting across experts, initialized from the multi-scale scaling in Eq.~\eqref{eq:multi_scale}. This preserves the always-on structure (no routing), can be absorbed into the low-rank factors (hence remains mergeable), and is used purely as a stabilization mechanism during training. In our experiments, unless stated otherwise, we adopt a LoRA+ optimizer setting that uses different effective learning rates for the LoRA factors~\citep{hayou2024lora+}; this is an optimizer-side choice and orthogonal to the SOS-LoRA parameterization.

\subsubsection{Orthogonal Input-Side Diversification}
\label{ssec:orthogonal}

To reduce redundancy and training-time interference among experts, we encourage diversity of input-side directions across $\{A_k\}$ through initialization and a lightweight regularizer.

\paragraph{Cross-expert orthogonal initialization with zero initial update.}
Assuming $r_{\text{tot}} = K r' \le m$, we initialize experts to span diverse input-side directions while keeping the initial update exactly zero. Concretely, we sample $G \in \mathbb{R}^{m \times (K r')}$ with i.i.d.\ Gaussian entries, compute a thin QR factorization $G = QR$ with $Q^\top Q = I_{K r'}$, and partition $Q$ into $K$ blocks $Q = [Q_1,\ldots,Q_K]$ with $Q_k \in \mathbb{R}^{m \times r'}$. We set $A_k = Q_k$ and initialize $B_k = 0$ for all $k$. This yields $\Delta W = \sum_{k=1}^K s_k A_k B_k = 0$ at initialization, preserving the backbone behavior at the start of fine-tuning.\footnote{If $r_{\text{tot}} > m$, one can fall back to per-expert orthogonalization (orthogonalizing each $A_k$ independently), which preserves within-expert orthogonality but cannot guarantee global cross-expert orthogonality.} Under this scheme, $B_k$ receives gradients immediately (Eq.~\eqref{eq:grad_scale}), while $\partial \mathcal{L} / \partial A_k = 0$ at initialization since $B_k = 0$. Thus, $A_k$ starts updating only after $B_k$ becomes non-zero, which stabilizes early training while starting from cross-expert-diverse input directions.

\paragraph{Cross-expert input-direction regularization.}
We regularize the \emph{input-side directions} by penalizing cross-expert correlations between the column spaces of $\{A_k\}$. Let $\tilde{A}_k \in \mathbb{R}^{m \times r'}$ denote the column-wise $\ell_2$-normalized version of $A_k$, i.e., $\tilde{a}=a/(\|a\|_2+\varepsilon)$ for each column $a$ and a small $\varepsilon>0$. We define
\begin{equation}
    \mathcal{L}_{\text{orth}}
    =
    \frac{\lambda}{r'^2} \sum_{1 \le k < \ell \le K}
    \left\|
        \tilde{A}_k^\top \tilde{A}_{\ell}
    \right\|_F^2,
    \label{eq:l_orth}
\end{equation}
which equals the sum of squared cosine similarities between all pairs of normalized columns across different experts. This directly discourages experts from collapsing to the same input-side directions while leaving within-expert structure unconstrained. In implementation, $\lambda$ can be scheduled (e.g., delayed ramp-up), and Eq.~\eqref{eq:l_orth} can be implemented either as an explicit loss term or via an equivalent gradient form without changing the underlying regularization objective.

\section{Experiments}
\label{sec:experiments}

\paragraph{Models and Datasets.}
We consider decoder-only LLMs (Llama 2-7B~\citep{llama2}, Llama 3-8B~\citep{llama3}) and an encoder-only model (RoBERTa-base~\citep{liu2019roberta}).
Our benchmarks cover three capability groups:
(i) reasoning and knowledge-intensive QA: BoolQ~\citep{clark2019boolq}, PIQA~\citep{bisk2020piqa}, SocialIQA~\citep{sap2019socialiqa}, HellaSwag~\citep{zellers2019hellaswag}, WinoGrande~\citep{sakaguchi2021winogrande}, ARC-E/ARC-C~\citep{clark2018think}, and OpenBookQA~\citep{mihaylov2018can};
(ii) natural language understanding: GLUE~\citep{wang2018glue};
and (iii) math reasoning: GSM8K~\citep{cobbe2021training} and MATH~\citep{hendrycks2021measuring}.

\paragraph{Compared Methods.}
We compare SOS-LoRA with full fine-tuning (Full FT) as a reference and standard LoRA~\citep{hu2022lora} as the primary PEFT baseline. We additionally include representative LoRA variants that appear in the main results tables: DoRA~\citep{liu2024dora} (decomposing weights into magnitude and direction, with low-rank adaptation on the directional component), AdaLoRA~\citep{zhang2023adalora} (adaptive rank allocation), DyLoRA~\citep{valipour2022dylora} (supporting a range of effective ranks), MELoRA~\citep{ren2024melora} (a mini-ensemble of low-rank adapters), PiSSA~\citep{meng2024pissa} (SVD-based initialization), Delta-LoRA~\citep{zi2023delta} (propagating updates via deltas of successive low-rank products), MixLoRA~\citep{li2024mixlora} (a routing-based sparse mixture of LoRA experts inserted into feed-forward blocks), and MSP-LoRA~\citep{zhao2025msplora} (a multi-scale pyramid LoRA with shared and layer-specific components). For the cost analysis in Table~\ref{tab:cost_gsm8k}, we report LoRA and Full FT as references; additional PEFT baselines are included where noted.

\paragraph{Implementation Details.}
All experiments are implemented in PyTorch with Hugging Face Transformers.
Unless otherwise noted, SOS-LoRA is applied to all linear projections:
for decoder-only LLMs (the Llama family), we adapt the attention projections (\texttt{q,k,v,o}) and MLP projections (\texttt{up,gate,down});
for RoBERTa-base, we adapt the self-attention and feed-forward linear layers.
Unless otherwise specified, we use a fixed total rank $r_{\text{tot}}=8$ per adapted matrix, decomposed into $K=4$ static experts ($r'=2$), with LoRA dropout rate 0.05.
We use fixed multi-scale scaling with $\gamma_{\max}=2.5$ (Eq.~\ref{eq:multi_scale}) and orthogonal initialization with cross-expert regularizer weight $\lambda=0.01$ (Eq.~\ref{eq:l_orth}).
We optimize with AdamW~\citep{loshchilov2019decoupled} and a linear learning-rate schedule with warmup ratio 0.1. Within each benchmark, we match the training data and training budget across methods.
For RoBERTa-base on GLUE, we train for 3 epochs with learning rate $2\times10^{-4}$ and batch size 32 (max sequence length 512).
Unless otherwise stated, Llama experiments use 3 epochs with learning rate $3\times10^{-5}$, per-device batch size 4 with 8 gradient accumulation steps (effective batch size 32), and max sequence length 2048.
For analyses that vary the rank budget or expert count (Figs.~\ref{fig:loss_dynamics} and \ref{fig:analysis_all}), we sweep $r_{\text{tot}}$ and/or $K$ while keeping all other settings fixed.
For the cost analysis in Table~\ref{tab:cost_gsm8k}, we fine-tune Llama 2-7B on the first 100k MetaMathQA samples~\citep{metamath} using DeepSpeed~\citep{rasley2020deepspeed} ZeRO-2 on 2$\times$ NVIDIA A40 (48GB), and report peak \emph{reserved} VRAM summed over the two devices together with end-to-end wall-clock time.
The Full FT accuracies follow the protocol described in the table note.
Unless stated otherwise, reported results are averaged over three independent runs with different random seeds; we report mean ± standard deviation where applicable.

\begin{table*}[t]
\centering
\caption{Main results on eight reasoning and knowledge-intensive benchmarks for the Llama family. We compare SOS-LoRA with strong PEFT baselines under a unified evaluation pipeline. Within each model block, the best and second-best PEFT results are marked in bold and underlined, respectively. We additionally report ChatGPT as a reference (see Appendix for evaluation details).}
\label{tab:llama_commonsense}
\setlength{\tabcolsep}{4.2pt}
\renewcommand{\arraystretch}{1.08}
\begin{tabular*}{\linewidth}{@{\extracolsep{\fill}}llccccccccc@{}}
\toprule
Model & Method & BoolQ & PIQA & SIQA & \makecell{Hella\\Swag} & \makecell{Wino\\Grande} & ARC-e & ARC-c & OBQA & Avg. \\
\midrule
\multicolumn{2}{l}{ChatGPT} & 73.1 & 85.4 & 68.5 & 78.5 & 66.1 & 89.8 & 79.9 & 74.8 & 77.0 \\
\midrule
\multirow{5}{*}{Llama 2-7B}
& LoRA  & 69.8 & 79.9 & 79.5 & 83.6 & 82.6 & 79.8 & 64.7 & 81.0 & 77.6 \\
& DoRA  & 71.8 & 83.7 & 76.0 & 89.1 & 82.6 & 83.7 & 68.2 & 82.4 & 79.7 \\
& MixLoRA & 72.0 & 82.8 & 77.3 & 92.9 & 75.7 & 77.4 & 57.5 & 81.1 & 77.1 \\
& PiSSA & \underline{75.0} & \underline{87.0} & \underline{81.6} & \underline{95.0} & \underline{86.5} & \underline{88.5} & \underline{75.9} & \underline{86.4} & \underline{84.5} \\
& \textbf{SOS-LoRA} & \textbf{75.7} & \textbf{87.5} & \textbf{83.2} & \textbf{95.8} & \textbf{87.8} & \textbf{89.2} & \textbf{76.9} & \textbf{88.1} & \textbf{85.5} \\
\addlinespace[2pt]
\midrule
\multirow{5}{*}{Llama 3-8B}
& LoRA  & 70.8 & 85.2 & 79.9 & 91.7 & 84.3 & 84.2 & 71.2 & 79.0 & 80.8 \\
& DoRA  & 74.6 & 89.3 & 79.9 & 95.5 & 85.6 & 90.5 & 80.4 & 85.8 & 85.2 \\
& MixLoRA & 74.6 & 87.2 & 78.1 & 93.0 & 81.4 & 86.1 & 79.3 & 84.0 & 83.0 \\
& PiSSA & \textbf{77.2} & \underline{90.0} & \underline{82.9} & \underline{96.6} & \underline{88.4} & \textbf{93.6} & \underline{82.4} & \underline{87.4} & \underline{87.3} \\
& \textbf{SOS-LoRA} & \underline{77.1} & \textbf{90.5} & \textbf{83.3} & \textbf{97.2} & \textbf{89.6} & \underline{93.4} & \textbf{83.9} & \textbf{89.2} & \textbf{88.0} \\
\bottomrule
\end{tabular*}
\end{table*}

We evaluate SOS-LoRA on eight reasoning and knowledge-intensive benchmarks. As shown in Table~\ref{tab:llama_commonsense}, SOS-LoRA achieves the highest average among the compared PEFT methods on both backbones. On Llama 2-7B, SOS-LoRA reaches 85.5 on average, improving over PiSSA (84.5) by 1.0 point and standard LoRA (77.6) by 7.9 points. SOS-LoRA also achieves the best result on all eight benchmarks in this setting, including PIQA (87.5 vs.\ 87.0 for PiSSA), and attains the best overall average. We also evaluate MixLoRA, a routing-based sparse MoE baseline, under the same pipeline: while it remains competitive on several benchmarks, its overall average is below both standard LoRA and SOS-LoRA (77.1 vs.\ 77.6 and 85.5, respectively). The trend persists on Llama 3-8B: SOS-LoRA attains an average of 88.0, surpassing PiSSA (87.3) by 0.7 and DoRA (85.2) by 2.8 points, while MixLoRA achieves 83.0. Overall, these results are consistent with SOS-LoRA's inductive bias: under a fixed rank budget, decomposing the update into static experts with pre-specified multi-scale scaling and encouraging cross-expert input-side direction diversity provides more effective adaptation than a single low-rank update.

\begin{table*}[t]
\centering
\caption{Performance comparison on the GLUE benchmark using RoBERTa-base. We evaluate SOS-LoRA against full fine-tuning (Full FT) and PEFT baselines under our unified training and evaluation pipeline. For each task, the best result is in bold and the second-best is underlined.}
\label{tab:glue_results_final}
\setlength{\tabcolsep}{4.5pt}
\renewcommand{\arraystretch}{1.08}
\begin{tabular*}{\linewidth}{@{\extracolsep{\fill}}lccccccccc@{}}
\toprule
Method & MRPC & RTE & CoLA & STS-B & SST-2 & QQP & QNLI & MNLI & Avg. \\
\midrule
\textit{Full FT}    & 88.2 & 84.1 & 64.6 & 90.6 & 94.3 & \textbf{92.0} & 92.7 & \underline{87.5} & 86.8 \\
\addlinespace[2pt]
\midrule
LoRA        & 89.9 & 85.9 & 62.4 & 91.4 & 94.4 & 90.8 & 92.6 & 86.9 & 86.8 \\
DyLoRA      & 89.5 & 84.5 & 61.1 & 91.1 & 94.3 & 90.2 & 92.2 & 86.3 & 86.2 \\
AdaLoRA     & 90.2 & 85.2 & 61.6 & 91.2 & 94.5 & 90.1 & 93.1 & 87.3 & 86.7 \\
Delta-LoRA  & 90.2 & \underline{87.0} & 63.8 & 91.6 & 95.1 & 90.9 & 93.1 & \underline{87.5} & \underline{87.5} \\
MELoRA      & \underline{90.9} & 86.6 & 64.1 & \underline{91.9} & \underline{95.4} & 90.8 & \underline{93.2} & 87.2 & \underline{87.5} \\
MSP-LoRA    & 90.1 & 80.8 & \underline{65.3} & 91.0 & 94.4 & 91.2 & 92.7 & 87.2 & 86.6 \\
\textbf{SOS-LoRA} & \textbf{91.7} & \textbf{88.1} & \textbf{65.6} & \textbf{92.4} & \textbf{95.7} & \underline{91.5} & \textbf{93.9} & \textbf{87.6} & \textbf{88.3} \\
\bottomrule
\end{tabular*}
\end{table*}

\subsection{Generalization to Encoder Architectures and NLU Tasks}
\label{ssec:glue_results}

To assess generalization to encoder architectures and NLU tasks, we evaluate SOS-LoRA on GLUE with RoBERTa-base~\citep{liu2019roberta}. As shown in Table~\ref{tab:glue_results_final}, SOS-LoRA achieves the highest average score (88.3) among the compared PEFT methods, improving over the strongest PEFT baselines under our evaluation pipeline (Delta-LoRA/MELoRA, 87.5) by 0.8 points, and also yielding a higher average than full fine-tuning in this setting (88.3 vs.\ 86.8). We additionally report MSP-LoRA~\citep{zhao2025msplora} as a recent multi-scale LoRA baseline, which attains 86.6 on average in our evaluation. SOS-LoRA achieves the best result on seven of the eight tasks, while Full FT is best on QQP, with consistent gains on paraphrase identification (MRPC), semantic similarity regression (STS-B), and natural language inference (RTE/QNLI/MNLI). Overall, these results indicate that decomposing a fixed low-rank budget into diversified static experts remains effective in encoder-based NLU settings.

\begin{figure}[t]
  \centering
  \includegraphics[width=\columnwidth]{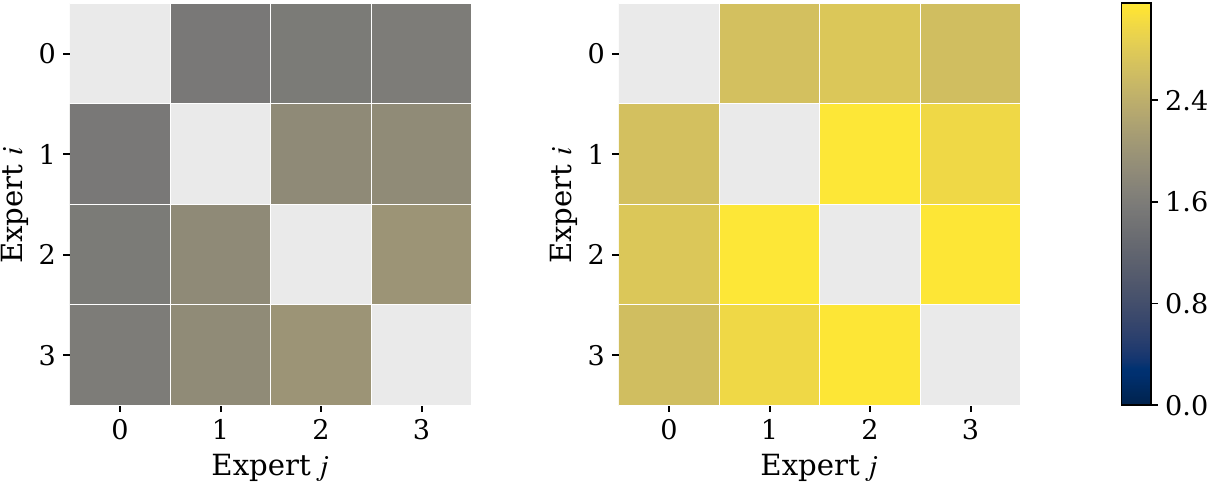}
  \caption{Orthogonal regularization reduces cross-expert similarity in input-side directions. Left: SOS-LoRA with the full method. Right: SOS-LoRA without $\mathcal{L}_{\text{orth}}$. Each entry is $\frac{1}{r'^2}\|\tilde A_k^\top \tilde A_\ell\|_F^2$, i.e., the mean squared cosine similarity between $\ell_2$-normalized input-side directions across experts ($K{=}4$), averaged over 224 adapted modules and scaled by $10^{6}$. Removing $\mathcal{L}_{\text{orth}}$ increases off-diagonal similarity.}
  \label{fig:expert_diversity_heatmap}
\end{figure}

\subsection{Cross-Expert Input-Side Direction Diversity Analysis}
\label{ssec:diversity_heatmap}

To assess whether SOS-LoRA learns diverse input-side directions, we analyze the $A$ factors after training. For each adapted linear module (224 in total), we compute a $K{\times}K$ cross-expert similarity matrix using the same squared-cosine measure as in Eq.~\ref{eq:l_orth}, namely $\frac{1}{r'^2}\|\tilde A_k^\top \tilde A_\ell\|_F^2$, where $\tilde A_k$ denotes column-wise $\ell_2$ normalization. As shown in Fig.~\ref{fig:expert_diversity_heatmap} (left vs.\ right), full SOS-LoRA exhibits lower off-diagonal similarity, whereas removing $\mathcal{L}_{\text{orth}}$ yields substantially higher off-diagonal values. This pattern is consistent with $\mathcal{L}_{\text{orth}}$ discouraging different experts from converging to redundant input-side directions and promoting better cross-expert separation.

\begin{figure}[t]
  \centering
  \includegraphics[width=\columnwidth]{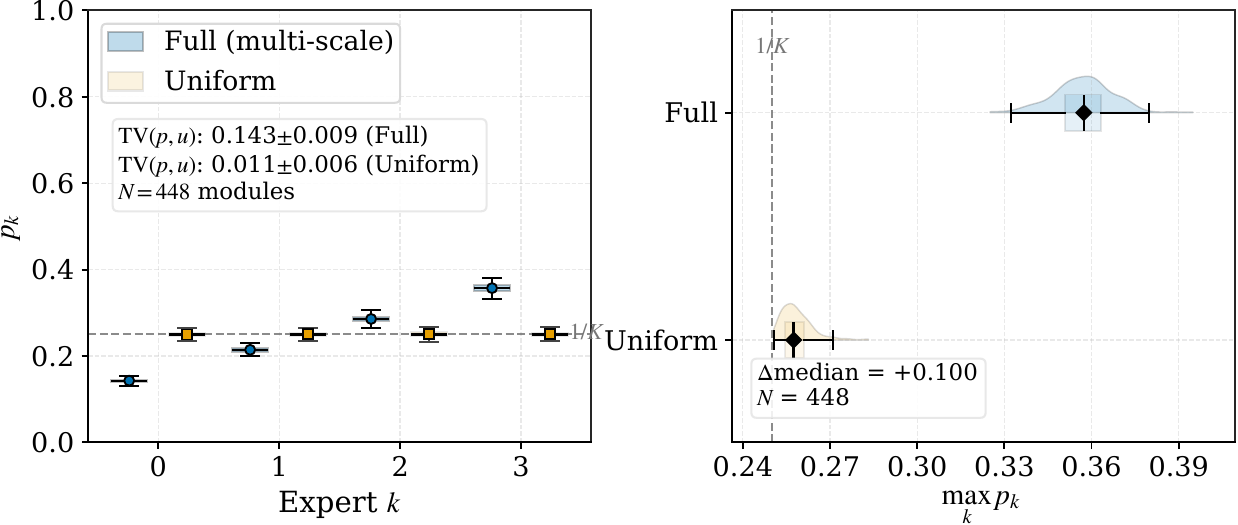}
  \caption{Multi-scale expert scaling leads to more separated per-expert update magnitudes. Left: per-expert relative update magnitude $p_k=\|\Delta W_k\|_F/\sum_{j=1}^{K}\|\Delta W_j\|_F$ across modules ($K{=}4$). Right: distribution of the top-1 share $\max_k p_k$, where multi-scale scaling yields a more concentrated contribution profile.}
  \label{fig:scale_separation}
\end{figure}

\subsection{Scale Separation Analysis}
\label{ssec:scale_separation}

We test whether the pre-specified multi-scale expert scaling promotes scale-separated update magnitudes across experts. Using Llama 2-7B, we compare \textsc{Full} SOS-LoRA ($\gamma_{\max}{=}2.5$) with a \textsc{Uniform} control ($\gamma_{\max}{=}1.0$) and measure each expert's relative update magnitude $p_k=\|\Delta W_k\|_F \big/ \sum_{j=1}^{K}\|\Delta W_j\|_F$. As shown in Fig.~\ref{fig:scale_separation}, the left panel indicates that \textsc{Full} yields a more clearly separated contribution profile across experts, whereas \textsc{Uniform} is closer to a balanced allocation around $1/K$. The right panel further shows that the top-1 share $\max_k p_k$ is consistently larger under \textsc{Full}. These results are consistent with multi-scale scaling separating update magnitudes across experts under a fixed rank budget, which may in turn encourage more differentiated expert roles during training.

\begin{figure}[t]
    \centering
    \includegraphics[width=\columnwidth]{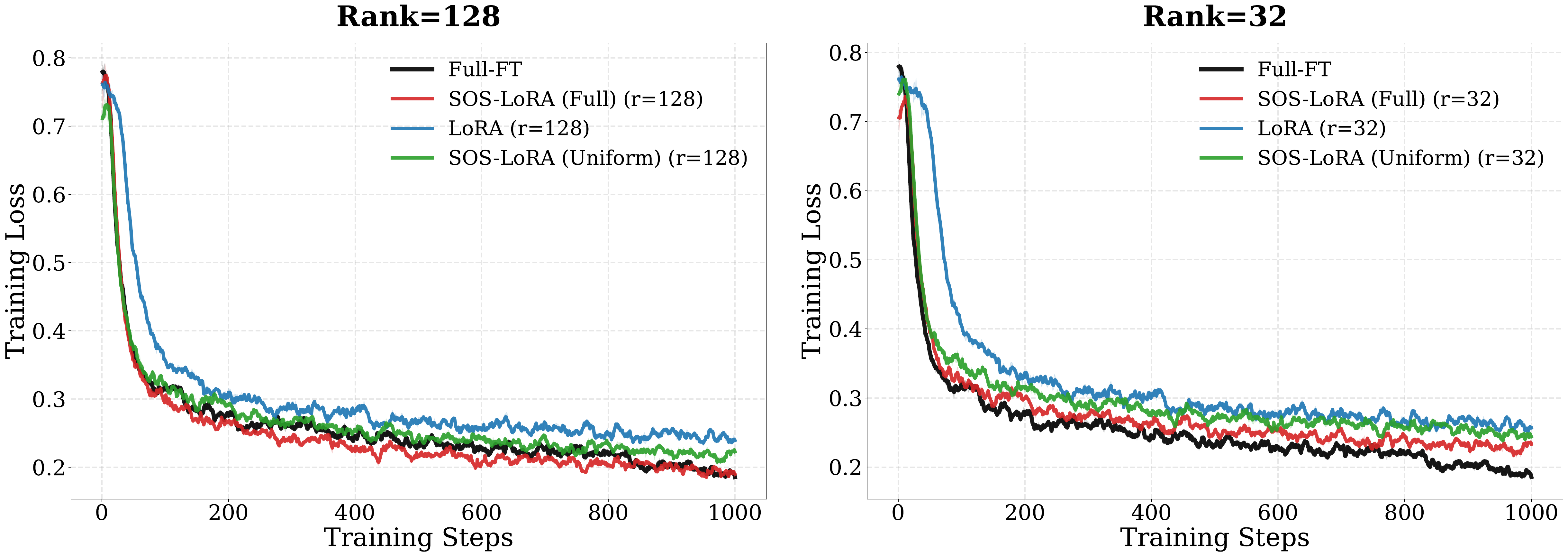}
    \caption{Optimization dynamics on GSM8K. Training loss comparing SOS-LoRA with standard LoRA and the uniform-scale ablation at two rank budgets ($r_{\text{tot}}{=}32$ and $r_{\text{tot}}{=}128$). SOS-LoRA exhibits faster early loss reduction and lower final loss.}
    \label{fig:loss_dynamics}
\end{figure}

\begin{figure*}[t]
    \centering
    \includegraphics[width=\textwidth]{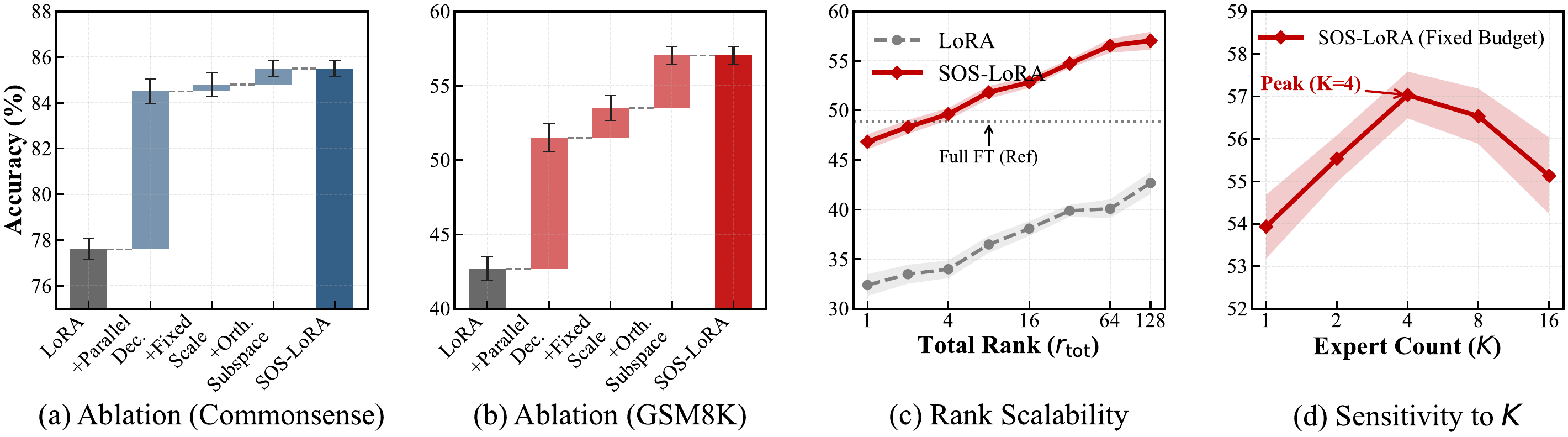}
    \caption{Analysis of SOS-LoRA on Llama 2-7B.
(a) Ablation on reasoning and knowledge-intensive benchmarks (standard budget $r_{\text{tot}}{=}8$): stepwise variants illustrate the contribution of each component.
(b) Ablation on GSM8K (high-rank budget $r_{\text{tot}}{=}128$): under our training setup, performance increases with rank and reaches 57.03\%.
(c) Rank scalability on GSM8K: SOS-LoRA consistently outperforms LoRA and matches or surpasses Full FT at higher ranks in our setup.
(d) Sensitivity to expert count $K$ on GSM8K (fixing $r_{\text{tot}}{=}128$): performance peaks at $K{=}4$, balancing cross-expert diversification and per-expert capacity.}
\label{fig:analysis_all}
\end{figure*}

\begin{table*}[t]
    \centering
    \small
    \setlength{\tabcolsep}{4.5pt}
    \renewcommand{\arraystretch}{1.15}
    \caption{Cost--accuracy trade-off on GSM8K and MATH when fine-tuning Llama 2-7B on the first 100k MetaMathQA samples using 2$\times$ NVIDIA A40 (48GB). Peak VRAM is the maximum \emph{reserved} memory.}
    \label{tab:cost_gsm8k}
    \begin{tabular*}{.8\linewidth}{@{\extracolsep{\fill}}lcccc@{}}
        \toprule
        Method & Peak VRAM (GB)$\downarrow$ & Time (wall)$\downarrow$ & GSM8K (\%)$\uparrow$ & MATH (\%)$\uparrow$ \\
        \midrule
        Full FT$^\dagger$ & N/A (OOM) & N/A & 48.90 & 7.48 \\
        \addlinespace[3pt]
        \midrule
        LoRA     & \textbf{66.32} & \textbf{4:31} & 42.68 & 5.92 \\
        SOS-LoRA & 67.45 & 4:52 & \textbf{57.03} & \textbf{10.06} \\
        \bottomrule
    \end{tabular*}

    \vspace{2pt}
    \begin{minipage}{.8\linewidth}
    \footnotesize \textit{$^\dagger$ Full FT is out of memory under the same 2$\times$ NVIDIA A40 (48GB) + DeepSpeed ZeRO-2 configuration. Its reported accuracies are obtained with a modified training recipe and are therefore not directly comparable in cost to LoRA and SOS-LoRA. See Appendix~A.7 for details.}
    \end{minipage}
\end{table*}

\subsection{Optimization Dynamics}
\label{ssec:optimization_dynamics}

We examine training dynamics in Fig.~\ref{fig:loss_dynamics} by fine-tuning Llama 2-7B on the first 100k MetaMathQA samples~\citep{metamath}. At both low ($r_{\text{tot}}{=}32$) and high ($r_{\text{tot}}{=}128$) rank budgets, SOS-LoRA shows faster early loss reduction and lower final loss than standard LoRA under the same training setup. The \textsc{Uniform} ablation reduces this advantage, suggesting that the pre-specified multi-scale scaling scheme contributes materially to the observed acceleration.

\subsection{Ablation and Sensitivity Analysis}
\label{ssec:analysis}

We analyze SOS-LoRA on Llama 2-7B to assess the contribution of each design component and sensitivity to key design choices and hyperparameters (Fig.~\ref{fig:analysis_all}). Under fixed rank budgets, the ablations in Fig.~\ref{fig:analysis_all}(a--b) show that parallel expert decomposition provides the largest gain, while multi-scale scaling and orthogonal regularization yield consistent additional improvements, indicating complementary benefits beyond decomposition alone. Fig.~\ref{fig:analysis_all}(c) further shows that SOS-LoRA improves over standard LoRA across a wide range of ranks ($r_{\text{tot}}\in[1,128]$) and matches or surpasses the Full FT baseline at higher ranks in our setup. Finally, Fig.~\ref{fig:analysis_all}(d) suggests that performance peaks around $K{=}4$ at fixed $r_{\text{tot}}{=}128$: larger $K$ reduces the per-expert rank and may limit per-expert capacity, whereas $K{=}1$ removes cross-expert diversification, consistent with a diversification--capacity trade-off.

\subsection{Computational Cost Analysis}
\label{sssec:cost_analysis}

We benchmark the cost--accuracy trade-off on Llama 2-7B using 2$\times$ NVIDIA A40 (48GB) GPUs with DeepSpeed ZeRO-2. We fine-tune on the first 100k MetaMathQA samples~\citep{metamath} and evaluate on GSM8K and MATH. Table~\ref{tab:cost_gsm8k} shows that SOS-LoRA achieves higher accuracy than standard LoRA on both benchmarks (57.03\% vs.\ 42.68\% on GSM8K; 10.06\% vs.\ 5.92\% on MATH). Relative to LoRA, SOS-LoRA improves by 14.35 points on GSM8K and 4.14 points on MATH. This gain comes with a modest increase in training cost (4:52 vs.\ 4:31 wall-clock time; 67.45 vs.\ 66.32 GB peak reserved VRAM), likely due primarily to the orthogonal regularizer. Full fine-tuning is not feasible under the same configuration because of OOM; its reported accuracies are obtained with a modified recipe and are therefore not directly comparable in cost to LoRA and SOS-LoRA (Table~\ref{tab:cost_gsm8k}$^\dagger$; see Appendix~A.7). Overall, SOS-LoRA offers a favorable accuracy--cost trade-off under this protocol.

\section{Conclusion}
\label{sec:conclusion}

We identify shared input-side coupling as a limitation of low-rank adaptation under a fixed rank budget, where a single input-side pathway can couple heterogeneous adaptation signals and induce optimization interference. We propose \textbf{Static Orthogonal Subspace LoRA (SOS-LoRA)}, which decomposes each low-rank update into $K$ \emph{static} (always-on, non-routed) experts with a fixed multi-scale scaling scheme and explicit input-side direction diversification via orthogonal initialization and a lightweight regularizer. SOS-LoRA is mergeable into the base model, adding no inference-time parameters or latency after merging. Experiments on both encoder-only (RoBERTa) and decoder-only (Llama 2/3) backbones show consistent gains over PEFT baselines under matched training budgets, suggesting that explicit input-side direction diversification can provide a practical route to stronger parameter-efficient adaptation.

\section*{Limitations}
\label{sec:limitations}

Our evaluation covers both encoder-only (RoBERTa-base) and decoder-only (Llama 2/3) backbones, but it is still limited to a relatively narrow range of model sizes and benchmark suites. Performance in larger models and broader settings (e.g., long-context, multilingual, and tool-use scenarios) remains untested. SOS-LoRA also introduces additional hyperparameters (e.g., expert count $K$, multi-scale range $\gamma_{\max}$, and regularization weight $\lambda$). Although we use a consistent default configuration and provide ablations and sensitivity analyses for key factors, the best settings may still vary with architecture, task, and training recipe. Finally, our analysis relies on geometric proxies (e.g., cross-expert squared-cosine similarity between input-side directions and relative update magnitudes) that capture redundancy and scale separation, but do not directly measure functional specialization. Complementary behavior-level analyses, such as targeted interventions or task-conditioned attribution/probing, would help strengthen these interpretations.

\section*{Acknowledgments}
This work is supported by the National Key Research and Development Program of China (No.2023YFF0905400), the National Natural Science Foundation of China (No.U2341229) and the Reform Commission Foundation of Jilin Province (No.2024C003).

\bibliography{references}

\newpage

\appendix

\section{Appendix}

\subsection{Method Details of SOS-LoRA}
\label{app:sos_details}

SOS-LoRA replaces a single rank-$r_{\text{tot}}$ LoRA update with $K$ \emph{static} (always-on, non-routed) low-rank experts of rank $r' = r_{\text{tot}}/K$. For a linear layer with frozen weight $W_0 \in \mathbb{R}^{m \times n}$ and input activations $X \in \mathbb{R}^{N \times m}$, the SOS-LoRA forward pass is
\begin{equation}
Y = XW_0 + \sum_{k=1}^{K} s_k \, X A_k B_k,
\end{equation}
where $A_k \in \mathbb{R}^{m \times r'}$, $B_k \in \mathbb{R}^{r' \times n}$, and $s_k > 0$ are fixed expert scales set by the multi-scale schedule in Eq.~\ref{eq:multi_scale}. The parameterization is mergeable: after training, the effective update is $\Delta W = \sum_{k=1}^{K} s_k A_k B_k$, which can be added to $W_0$ to obtain a standard dense weight for inference, incurring no additional inference-time parameters or latency after merging. We use cross-expert orthogonal initialization with $B_k = 0$, so that $\Delta W = 0$ at initialization while the $A_k$ start from diverse input-side directions. The cross-expert diversity regularizer is the squared-cosine penalty in Eq.~\ref{eq:l_orth}.

\subsection{Algorithmic Summary of SOS-LoRA}
\label{app:sos_algorithm}
For clarity and reproducibility, we provide a high-level outline of SOS-LoRA for a single linear layer. The same procedure is applied to every adapted linear projection while keeping the pretrained weights frozen; all steps correspond to the definitions in Section~\ref{sec:methodology}. The procedure is summarized in Algorithm~\ref{alg:soslora}.

\begin{algorithm}[t]
\caption{SOS-LoRA training and merging for a single linear layer}
\label{alg:soslora}
\begin{algorithmic}[1]
\Require Frozen $W_0\in\mathbb{R}^{m\times n}$; total rank $r_{\text{tot}}$; number of experts $K$ ($r'=r_{\text{tot}}/K$); fixed scales $\{s_k\}$ (Eq.~\ref{eq:multi_scale})
\State \textbf{Initialize:} when $r_{\text{tot}}=Kr' \le m$, sample $G\in\mathbb{R}^{m\times Kr'}$, compute thin QR $G=QR$, and split $Q=[Q_1,\ldots,Q_K]$
\State \hspace{1.35em} Set $A_k\gets Q_k$ and $B_k\gets 0$ for $k=1,\ldots,K$ \Comment{$\Delta W=0$ at initialization}
\For{each training step}
    \State \textbf{Forward:} $Y \gets XW_0 + \sum_{k=1}^{K} s_k\, X A_k B_k$
    \State \textbf{Objective:} $\mathcal{L}\gets \mathcal{L}_{\text{task}}(Y) + \mathcal{L}_{\text{orth}}$ \Comment{$\mathcal{L}_{\text{orth}}$ is defined in Eq.~\ref{eq:l_orth}}
    \State \textbf{Update:} update $\{A_k,B_k\}_{k=1}^{K}$ with the chosen optimizer; keep $W_0$ frozen
\EndFor
\State \textbf{Merge for inference:} $\Delta W\gets \sum_{k=1}^{K} s_k A_kB_k$; $W\gets W_0+\Delta W$; discard adapter parameters
\end{algorithmic}
\end{algorithm}

\subsection{Training Setup and Hyperparameters}
\label{app:train_setup}

All experiments are implemented in PyTorch with Hugging Face Transformers and the PEFT library. Unless otherwise noted, SOS-LoRA is applied to all linear projections. For decoder-only LLMs (the Llama family), we adapt the attention projections (\texttt{q,k,v,o}) and MLP projections (\texttt{up,gate,down}); for RoBERTa-base, we adapt the self-attention and feed-forward linear layers. Unless otherwise specified, we use $r_{\text{tot}}{=}8$ per adapted matrix with $K{=}4$ experts ($r'{=}2$), LoRA dropout rate 0.05, $\gamma_{\max}{=}2.5$ in Eq.~\ref{eq:multi_scale}, and $\lambda{=}0.01$ in Eq.~\ref{eq:l_orth}. We optimize with AdamW with a linear learning-rate schedule and warmup ratio 0.1. For RoBERTa-base on GLUE, we train for 3 epochs with learning rate $2\times 10^{-4}$, batch size 32, and max sequence length 512. Unless otherwise stated, Llama experiments use 3 epochs with learning rate $3\times 10^{-5}$, per-device batch size 4 with 8 gradient accumulation steps (effective batch size 32), and max sequence length 2048. Unless stated otherwise, reported results are averaged over three independent runs with different random seeds; we report mean $\pm$ standard deviation where applicable. For sweeps over $r_{\text{tot}}$ and/or $K$, we keep all other settings fixed to isolate the effect of varying rank capacity and expert granularity.

\subsection{Datasets, Splits, and Metrics}
\label{app:data_metrics}

We use public benchmarks with their official splits. Reasoning and knowledge-intensive evaluation includes BoolQ, PIQA, SocialIQA, HellaSwag, WinoGrande, ARC-E/ARC-C, and OpenBookQA, evaluated under a unified prompting and scoring pipeline with accuracy as the metric. NLU evaluation uses GLUE with task-specific metrics following the benchmark conventions~\citep{wang2018glue}: CoLA uses Matthews correlation, STS-B uses correlation, SST-2/QNLI/RTE use accuracy, and MRPC/QQP are conventionally reported with both F1 and accuracy. In Table~\ref{tab:glue_results_final}, we report accuracy for MRPC and QQP (while retaining F1 in our released logs/configs), report MNLI matched accuracy (MNLI-m), and report the simple average over the eight tasks shown (excluding WNLI). Math reasoning evaluation uses GSM8K and MATH: GSM8K is evaluated by exact-match accuracy on the final answer, and MATH is evaluated by final-answer accuracy under the same post-processing rules across methods~\citep{cobbe2021training,hendrycks2021measuring}.

\subsection{Baselines and Configuration Matching}
\label{app:baselines}

We compare SOS-LoRA against Full FT as a reference and LoRA as the primary PEFT baseline. We additionally include representative LoRA variants reported in the main results tables: DoRA, AdaLoRA, DyLoRA, MELoRA, PiSSA, Delta-LoRA, MixLoRA, and MSP-LoRA. For fair comparison, we match (i) backbone checkpoints, (ii) training data and the number of steps/epochs, (iii) the set of adapted modules, unless a baseline by design constrains insertion locations, and (iv) the total low-rank budget per adapted matrix whenever a direct rank match is supported. When a baseline introduces additional components (e.g., routing or extra parameters), we run it under the same training pipeline and overall optimization budget, while following the authors' recommended defaults for baseline-specific hyperparameters. All configurations are recorded in our released files.

\subsection{Mechanistic Analyses: Definitions and Computation}
\label{app:analysis_defs}

\textbf{Cross-expert input-side direction similarity.}
For each adapted module, we compute a $K\times K$ matrix with entries $\frac{1}{r'^2}\|\tilde A_k^\top \tilde A_\ell\|_F^2$, where $\tilde A_k$ is obtained by $\ell_2$-normalizing each column of $A_k$. This quantity equals the mean squared cosine similarity over all cross-expert column pairs and matches the quantity used in the regularizer in Eq.~\ref{eq:l_orth}. We report both module-wise matrices and their average across all adapted modules (224 for Llama 2-7B under our adaptation scope), and visualize them as heatmaps (Fig.~\ref{fig:expert_diversity_heatmap}).

\textbf{Per-expert relative update magnitude.}
For each adapted module, we compute $\Delta W_k = s_k A_kB_k$ and define $p_k=\|\Delta W_k\|_F\big/\sum_{j=1}^{K}\|\Delta W_j\|_F$. We plot the distribution of $p_k$ across modules together with the distribution of $\max_k p_k$ to characterize scale separation (Fig.~\ref{fig:scale_separation}).

\textbf{Loss dynamics.}
For the training curves in Fig.~\ref{fig:loss_dynamics}, we fine-tune Llama 2-7B on the first 100k MetaMathQA samples and log training loss at fixed intervals under identical optimization settings across methods, using the same token budget and sequence length. Curves show the mean across seeds.

\subsection{Efficiency Measurement Protocol}
\label{app:efficiency}

For the cost--accuracy study (Table~\ref{tab:cost_gsm8k}), we fine-tune Llama 2-7B on the first 100k MetaMathQA samples using 2$\times$ NVIDIA A40 (48GB) GPUs with DeepSpeed ZeRO-2. Peak VRAM is measured as the maximum \emph{reserved} memory (e.g., \texttt{torch.cuda.max\_memory\_reserved}) summed over the two devices during training. Wall-clock time is measured end-to-end under the same training budget and logging frequency across compared methods. Full FT is marked with $^\dagger$ because it requires a modified recipe to avoid OOM; we therefore report its accuracies only as a reference, and its cost is not directly comparable under the same configuration.

\subsection{ChatGPT Reference Evaluation}
\label{app:chatgpt_eval}

We include ChatGPT in Table~\ref{tab:llama_commonsense} only as a reference. We evaluate it under the same prompting and scoring format used in our unified evaluation pipeline, with fixed prompt templates and deterministic decoding when supported. The exact prompts, scripts, and post-processing rules are included in our release to enable reproduction of the evaluation procedure, subject to model/version changes by the provider. We do not use ChatGPT outputs for training, validation, or hyperparameter selection.

\subsection{Reproducibility Statement}
\label{app:reproducibility}

We provide the public code repository, training/evaluation scripts, and SOS-LoRA adapter checkpoints under an Apache 2.0 license. Our implementation is based on PyTorch, Hugging Face Transformers, and PEFT; experiments additionally use DeepSpeed for the ZeRO-2 efficiency study. We use publicly available pretrained checkpoints (e.g., Llama 2/3 and RoBERTa-base) and standard benchmarks (e.g., GLUE, GSM8K, MATH) with official splits. Section~\ref{sec:experiments} and Appendix~\ref{app:train_setup} report key hyperparameters, including adaptation scope, ranks, $K$, scaling, regularization, learning rates, batch sizes, sequence lengths, epochs, and optimizer schedule. Unless stated otherwise, all reported numbers are mean $\pm$ standard deviation over three independent runs with different random seeds.

\subsection{LLM Usage Statement}
\label{app:llm_usage}

We used a large language model (LLM) only for language polishing and proofreading of the manuscript text. The LLM was not used to design SOS-LoRA, develop algorithms, generate experimental results, or make methodological or scientific decisions. All technical contributions, experiments, and conclusions are our own, and we take full responsibility for the content of this paper.

\subsection{Responsible NLP Research Checklist}
\label{app:responsible_nlp}

We follow the ACL community's Responsible NLP Research guidelines and checklist and include the completed checklist in the final version, as required by the venue. Our experiments are conducted on widely used public benchmarks and standard model checkpoints; we report averaged results over multiple seeds and provide implementation details to support reproducibility. Although our work focuses on parameter-efficient adaptation rather than new data collection, downstream deployment of fine-tuned models should still consider dataset biases, evaluation limitations, and potential misuse, consistent with the checklist guidance.

\end{document}